\documentclass[conference,compsoc]{IEEEtran}
\usepackage{hyperref}
\usepackage{xcolor}

\usepackage{soul}
\usepackage[many]{tcolorbox}
\usepackage{enumitem}
\usepackage{caption}
\usepackage{inconsolata}
\tcbuselibrary{breakable}
\tcbuselibrary{skins, breakable}
\usepackage{graphicx}

\usepackage{xcolor}
\definecolor{TrustBlue}{RGB}{33,113,181}
\definecolor{main}{HTML}{5989cf}    
\definecolor{sub}{HTML}{cde4ff}     

\usepackage[many]{tcolorbox}

\newcommand{\cmark}{\textcolor{main}{\ding{51}}}

\ifCLASSOPTIONcompsoc
  \usepackage[nocompress]{cite}
\else
  \usepackage{cite}
\fi
\ifCLASSINFOpdf
\else
\fi

\usepackage{pifont}
\usepackage{xcolor}
\usepackage{fontawesome5}

\definecolor{contribbg}{RGB}{245,248,255}
\definecolor{contribbar}{RGB}{40,95,180}

\definecolor{takebg}{RGB}{232,242,255}
\definecolor{takebar}{RGB}{60,110,200}

\newtcolorbox{contribution}{
  enhanced,
  breakable,
  colback=contribbg,
  frame hidden,
  borderline west={4pt}{0pt}{contribbar},
  boxsep=4pt,
  left=8pt,
  right=6pt,
  top=5pt,
  bottom=5pt,
  before upper={
    \textcolor{contribbar}{\ding{72}}\hspace{6pt}
  },
  fontupper=\normalsize
}

\newtcolorbox{takeaway}{
  enhanced,
  breakable,
  colback=takebg,
  frame hidden,
  borderline west={4pt}{0pt}{takebar},
  boxsep=4pt,
  left=8pt,
  right=6pt,
  top=5pt,
  bottom=5pt,
  before upper={
    \textcolor{takebar}{\faLightbulb}\hspace{6pt}
  },
  fontupper=\normalsize
}

\newtcolorbox{action}{
  enhanced,
  breakable,
  colback=takebg, 
  frame hidden,
  borderline west={4pt}{0pt}{takebar},
  boxsep=4pt,
  left=8pt,
  right=6pt,
  top=5pt,
  bottom=5pt,
  before upper={
    \textcolor{takebar}{\faBullhorn}\hspace{6pt}\textbf{Call for Action:}\hspace{6pt}
  },
  fontupper=\normalsize
}

\usepackage{booktabs}   
\usepackage{tabularx}   
\usepackage[table]{xcolor} 

\begin{document}
%
\title{Inspectable AI for Science: A Research Object Approach to Generative AI Governance}

\author{\IEEEauthorblockN{Anonymous Authors}}
\author{\IEEEauthorblockN{Ruta Binkyte}
\IEEEauthorblockA{CISPA, Helmholtz Center \\for Information Security\\
}
\and
\IEEEauthorblockN{Sharif Abuaddba }
\IEEEauthorblockA{DATA61,CSIRO\\
}
\and
\IEEEauthorblockN{ Chamikara Mahawaga\\ Arachchige}
\IEEEauthorblockA{Data61, CSIRO\\
}
\and
\IEEEauthorblockN{Ming Ding}
\IEEEauthorblockA{Data61, CSIRO\\
}
\and
\IEEEauthorblockN{Natasha Fernandes }
\IEEEauthorblockA{Macquarie University\\
}
\and

\IEEEauthorblockN{Mario Fritz }
\IEEEauthorblockA{CISPA, Helmholtz Center \\for Information Security\\
}
}


%

\newcommand{\sharif}[1]{\textsf{\color{blue}{[{Sharif: #1}]}}}



\maketitle

\begin{abstract}
This paper introduces AI as a Research Object (AI-RO), a paradigm for governing the use of generative AI in scientific research. 
Instead of debating whether AI is an author or merely a tool, we propose treating AI interactions as structured, inspectable 
components of the research process. Under this view, the legitimacy of an AI-assisted scientific paper depends on how model 
use is integrated into the workflow, documented, and made accountable. Drawing on Research Object theory and FAIR principles, we propose a framework for recording model configuration, prompts, and outputs through interaction logs and metadata packaging. These properties are particularly consequential in security and privacy (S\&P) research, where provenance artifacts must satisfy confidentiality constraints, integrity guarantees, and auditability requirements that generic disclosure practices do not address. We 
implement a lightweight writing pipeline in which a language model synthesizes human-authored structured literature review notes under explicit constraints and produces a verifiable provenance record. 
We present this work as a position supported by an initial demonstrative workflow, arguing that governance of generative AI in science can be implemented as structured documentation, 
controlled disclosure, and integrity-preserving provenance capture. Based on this example, we outline and motivate a set of necessary 
future developments required to make such practices practical and widely adoptable.
\end{abstract}


%
\IEEEpeerreviewmaketitle
\section{Introduction}
The rapid proliferation of generative artificial intelligence (AI), particularly large language models (LLMs), has fundamentally transformed scientific writing and knowledge production. Across disciplines, researchers increasingly rely on AI systems for ideation, literature review, code generation, data analysis, manuscript drafting, and peer review~\cite{liang2024mapping}. This transformation is reflected in the growing number of AI-assisted submissions to academic venues and in emerging institutional guidelines governing AI use in research~\cite{editorials2023tools,thorp2023chatgpt,watkins2024guidance}

\paragraph{\textbf{Emerging Reality of AI Use in Science}} At the same time, this shift has introduced substantial uncertainty regarding authorship, originality, plagiarism, and intellectual accountability~\cite{altmae2023artificial}. A recent study by~\cite{ansari2026compounddeceptionelitepeer} identified 100 non-existent references in recently accepted papers at  NeurIPS'25, one of the major machine learning conferences, illustrating how AI-mediated workflows can undermine established norms of scholarly verification.  Such incidents highlight broader risks, including fully fabricated studies generated by AI, that challenge traditional expectations of academic responsibility~\cite{van2023ai}. In response, many journals and conferences now require authors to disclose AI assistance. 
Yet current disclosure practices remain largely self-attested and describe AI use in coarse categories, such as writing or grammar support, without mechanisms to verify the accuracy, scope, or procedural context of those claims~\cite{ieee_pspb_opsmanual,ganjavi2023bibliometric,taylorandfrancis_ai_policy_2026,elsevier_generative_ai_policies_2026 }.
   Statements such as “an LLM was used for language editing” reveal little about how AI shaped the document, whether confidentiality constraints were respected, or how claims of responsible use might be audited without exposing sensitive materials.  Transparency and confidentiality trade-off is particularly acute in security and privacy (S\&P) research. S\&P workflows routinely handle sensitive materials, such as vulnerability disclosures or adversarial test cases. Yet current disclosure mechanisms offer no way to verify whether confidentiality was respected during AI-assisted drafting, or to audit the procedural basis of AI-shaped claims without exposing 
sensitive materials. 

Finally, disclosure itself may be stigmatized, discouraging openness and potentially eroding trust rather than reinforcing it~\cite{schilke2025transparency}. We argue that the extent of automation will grow in the future,  indicated by developments in autonomous AI scientists and co-scientists~\cite{lu2024ai,gottweis2025towards}.  
Therefore, maintaining control and scientific integrity calls for urgent advances in establishing standardized,  verifiable, and practical tools for disclosing AI use in scientific research.

\paragraph{\textbf{Theoretical Approaches to the Use of AI}} Theoretical discussions of AI in scientific and creative work largely frame governance in terms of authorship: whether AI should be treated as an author, collaborator, or a tool. A prominent perspective conceptualizes AI as an instrument that extends human capabilities without possessing creative agency~\cite{sarkar2023enough}, aligning with legal positions that prohibit listing AI systems as authors of copyrighted materials~\cite{core2024louisiana,hesman2004machine}. Other scholars argue that AI may function as a legitimate collaborator when it substantively shapes intellectual direction, raising questions about credit and responsibility~\cite{dempere2025ai}. A third line of work reframes authorship around creative control, emphasizing human steering of prompts, curation, and iterative refinement~\cite{liu2025authorship}, while related research highlights psychological dimensions of ownership and engagement~\cite{fritz2025understanding,hwang2025owns}. Accountability-centered perspectives further stress that authorship entails responsibility for claims and ethical implications~\cite{bonadio2025copyrightability}. For example, the Committee on Publication Ethics (COPE) states that AI tools cannot meet authorship criteria because they cannot assume responsibility for scholarly claims~\cite{cope2023authorshipAI}. 

\paragraph{\textbf{Limitations of Existing Approaches}} Current theoretical and governance efforts typically rely on questions of authorship and contribution, disclosure statements, post-hoc AI detection efforts, or ad hoc prohibitions. Narrative disclosures are difficult to verify and often lack procedural detail. Detection methods are inherently error-prone and risk creating adversarial dynamics that discourage transparency. Blanket prohibitions ignore legitimate and beneficial uses such as language refinement or structured ideation, pushing AI use underground rather than governing it. Collectively, these approaches fail to provide structured, inspectable mechanisms for accountability and instead incentivize opacity over responsible transparency.

\paragraph{\textbf{AI as a Research Object}}  We therefore propose a shift from authorship-centered debates toward infrastructural accountability, that addresses not 
only transparency but the confidentiality, integrity, and auditability requirements that S\&P research demands. Drawing on the concept of Research Objects (ROs), structured digital artifacts that encapsulate datasets, software, workflows, and contextual metadata to support transparency and reproducibility~\cite{belhajjame2012research}, we argue that generative AI systems used in scientific writing should be governed as AI Research Objects (AI-ROs).  Concretely, AI-RO denotes a container of structured artifacts, including interaction logs, prompts, outputs, and metadata, which together capture the role of AI within a research workflow. Under this framing, AI is neither an author nor a collaborator, but a documented component of a research workflow whose provenance and configuration can be inspected alongside other scientific artifacts. 

In this work, we propose reframing the use of AI in scientific research through the lens of the research object framework, emphasizing key distinctions from traditional research objects such as datasets or code. We further present an illustrative end-to-end transparency pipeline centered on one of the most prevalent and contested applications of AI: literature review writing. We emphasize that this example is intended to demonstrate the feasibility of the approach and to stimulate discussion on concrete implementations and best practices, rather than to serve as a definitive solution. Finally, we outline potential next steps toward AI governance in scientific writing and peer review, and discuss current limitations and areas requiring further development.
\paragraph{\textbf{Scope}} This work presents a conceptual and infrastructural framework for governing AI-assisted scientific workflows rather than a finalized technical standard or empirical evaluation. Our demonstration focuses on writing a literature review as a concrete case study to illustrate feasibility and design principles. However, this paper is a position and infrastructure proposal intended to stimulate discussion rather than report an evaluated system. 

\begin{contribution}
\textbf{Contributions.}
1) We introduce AI-RO, a framework for governing generative AI
as inspectable research infrastructure; 2) We discuss operationalization of AI-RO; 3) We provide a feasibility demo; 4) We discuss the extension to governance interfaces; 5) Finally, we discuss limitations and provide actionable items for future development.
\end{contribution}
 The code and artifacts related to the paper are available at~\href{https://github.com/RutaBinkyte/AI-RO}{\texttt{https://github.com/RutaBinkyte/AI-RO}}, and~\href{https://osf.io/zbrvu}{\texttt{https://osf.io/zbrvu}}.

\section{Conceptual Foundations: Research Objects}

The notion of ROs was introduced as an abstraction for communicating, sharing, and reusing research results beyond traditional research papers~\cite{belhajjame2012research}. Rather than treating scientific output as a static document, the RO paradigm frames research as a structured digital artifact that encapsulates complementary components, including hypotheses, experimental workflows, datasets, software tools, and derived results. By packaging these elements together, Research Objects make the procedural context of scientific claims explicit, supporting transparency, reuse, and reproducibility in digitally mediated research.

Building on this foundation, a growing ecosystem of standards and infrastructures has emerged to operationalize ROs in practice. RO-Crate, for example, provides a lightweight mechanism for describing research artifacts using schema.org annotations expressed in JSON-LD, lowering the barrier to adopting structured metadata in everyday workflows~\cite{soiland2022creating}. In domains with heightened privacy or regulatory constraints, Trusted Research Environments (TREs) establish secure contexts in which data, tools, and workflows can be accessed and audited under formal governance structures~\cite{soiland2024five}. At a broader level, the FAIR principles articulate widely accepted expectations for scientific artifacts, namely that they should be Findable, Accessible, Interoperable, and Reusable~\cite{wilkinson2016fair}. Together, these developments reflect a shift toward treating computational artifacts as first-class objects of scientific accountability.

The next section builds on this conceptual foundation to formalize generative AI systems as AI-ROs, outlining how their configuration, interaction traces, and outputs can be governed using established infrastructural principles for scientific transparency and reproducibility.

\begin{takeaway}
\textbf{Takeaway 1.}
Research Objects reframe scientific outputs as structured, inspectable artifacts that preserve procedural context, enabling transparency, reproducibility, and accountability in computational research.
\end{takeaway}
\section{AI as Research Object}

Generative AI systems increasingly resemble established categories of scientific artefacts that are already governed as Research Objects. Like datasets, they embed prior information that conditions future results. Like software libraries, they implement reusable computational procedures. Like workflows, they transform inputs into outputs through configurable processes. In all these respects, AI systems function as components of research infrastructure rather than as independent epistemic agents. Their scientific relevance and/or risks arise from how they are integrated into scientific workflows.

 In addition, similarly to traditional ROs, AI systems have identifiable versions, parameter settings (e.g., temperature), execution contexts, and interaction histories.  This allows AI influence on outcomes to be documented and inspected. Therefore, treating AI usage as a Research Object allows researchers to leverage familiar tools and standards to support transparency, reproducibility, and auditability.

At the same time, AI Research Objects introduce properties that distinguish them from traditional artefacts and raise new methodological considerations. First, generative AI systems are inherently stochastic, meaning that identical inputs may yield different outputs even under controlled conditions. Second, many models are proprietary, limiting access to internal parameters, training data, or execution details that would traditionally support reproducibility. Third, AI-generated content often resembles human cognitive production, creating social and epistemic expectations that do not arise for conventional software or datasets. These characteristics complicate inspection and accountability, but they do not invalidate the Research Object framing. Instead, they motivate the use of additional documentation mechanisms, such as interaction logs and structured user prompts.

Table~\ref{tab:ro-comparison} summarizes the relationship between traditional Research Objects and AI Research Objects across key governance dimensions. Both  ROs and AI-ROs are digital artefacts embedded in scientific workflows, subject to versioning, documentation, and human accountability. However, differences lie in the granularity of configuration, the limits of reproducibility, and the need to capture interaction-level provenance. For AI systems, parameters such as sampling settings and prompt histories become part of the methodological record, while evaluation shifts toward assessing output quality under human oversight. 

\begin{table}[t]
\centering
\footnotesize
\begin{tabular}{p{0.08\textwidth} p{0.15\textwidth} p{0.15\textwidth}}
\hline
\textbf{Governance Dimension} & \textbf{Traditional Research Objects} & \textbf{AI Research Objects} \\
\hline
Ontological status & Digital scientific artefact & Digital scientific artefact \\
Typical examples & Datasets, software, workflows & AI use logs, AI agents \\
Versioning & Dataset/software versions & Model versions, checkpoints \\
Configuration & Parameters, scripts & Temperature, Top-p, Max tokens \\
Provenance & Data sources, pipelines & Prompt logs \\
Reproducibility & Full: Re-run workflow & Limited: Re-run model with same context \\
Documentation & README, metadata & AI-RO Inspection Cards, interaction logs \\
Evaluation & Performance metrics & Output quality, human oversight \\
Accountability & Human researchers & Human researchers \\
Automation & Low & Medium to High (in agentic settings) \\
\hline
\end{tabular}
\caption{Systematic comparison between traditional Research Objects and AI Research Objects}
\label{tab:ro-comparison}
\end{table}

\begin{takeaway}
\textbf{Takeaway 2.}
 AI-ROs share the structure and workflow integration of traditional ROs while introducing stochastic and interaction-driven behavior.
\end{takeaway}

\begin{figure}[t]
\centering
\footnotesize
\begin{tcolorbox}[
  colback=blue!3,
  colframe=blue!60!black,
  title=\textbf{AI Research Object Inspection Card (AI-ROIC)},
  fonttitle=\bfseries,
  boxrule=1pt,
  arc=2mm,
  width=\linewidth,
  left=3mm,
  right=3mm,
  top=2mm,
  bottom=2mm
]

\textbf{Run ID:} ro-background

\textbf{Research Topic:} Paradigms of authorship in generative AI creation

\medskip
\textbf{Model Configuration}
\begin{itemize}[leftmargin=5mm]
\item Interface: OpenAI-compatible API
\item Model class: LLaMA 3.1 (8B parameter family)
\item Temperature: 0.2
\item Top-p: 1.0
\item Max tokens: 1200
\item Input bundle SHA-256: \texttt{79b576eacbeee64171\\dfe0bc8cd7a6e5e09da7f25259a97819cb\\ac5ce35d0860}
\end{itemize}

\medskip
\textbf{Artifacts Released}
\begin{itemize}[leftmargin=5mm]
\item Privatized interaction logs (taxonomy + synthesis)
\item Generated taxonomy structure (JSON)
\item Draft synthesis text (Markdown)
\item Audit table (CSV)
\end{itemize}

\medskip
\textbf{Intended Use}

Assistive structural synthesis and drafting using human-authored inputs. The system was not treated as an autonomous factual authority.

\medskip
\textbf{Human Oversight}

All interpretations, claims, and references were verified by the authors.

\medskip
\textbf{Disclosure}

A generative language model was used for structural synthesis and drafting under constrained prompts. Prompts and intermediate artifacts are released to support transparency.

\medskip
\textbf{Limitations}

Outputs may contain model biases or structural artifacts. Human validation was required.

\medskip
\textbf{Reproducibility Note}

Exact regeneration may vary due to backend nondeterminism. Parameters and artifact hashes support approximate replication.

\end{tcolorbox}

\caption{Illustrative  AI Research Object Inspection Card (AI-ROIC)} summarizing configuration, artifacts, oversight, and reproducibility information for the generative assistance workflow.
\label{fig:model-card}

\end{figure}

\section{Operationalizing AI Research Objects}

Treating AI systems as Research Objects requires practical mechanisms for documentation, storage, inspection, and reuse. An operational AI Research Object should, at a minimum, capture the model identifier and version, system configuration and parameters, prompt and interaction logs, generated outputs, and human editing traces that distinguish AI-produced content from final authorial decisions. These artifacts form a provenance record that documents how AI systems influenced the research process. In alignment with FAIR principles, such packages can be indexed, stored, and shared using persistent identifiers and interoperable metadata schemas, enabling downstream inspection and reuse without redefining existing scientific infrastructure~\cite{wilkinson2016fair}.


\paragraph{\textbf{AI Agent as RO}} This operational perspective is particularly important when considering emerging agentic AI systems. Recent work demonstrates how AI agents can act as interactive interfaces to scientific content, encapsulating explanation, synthesis, or workflow execution within executable artifacts~\cite{miao2025paper2agent}. More advanced ``AI scientist" systems aim to automate hypothesis generation, experimental design, and analysis pipelines. While such systems promise efficiency, they also risk obscuring the procedural basis of scientific claims if their internal decision processes remain opaque. Framing agents as AI Research Objects provides a governance mechanism.  AI-ROs should capture both the agent system (configuration, orchestration) and its execution traces and outputs. As automation deepens, the importance of structured provenance will only grow: both for preserving the control and integrity of scientific research, and for using these artifacts to advance further development of scientific AI agents. 

\begin{takeaway}
\textbf{Takeaway 3.}
Treating AI and AI agents as Research Objects turns interaction histories into auditable artifacts, preserving transparency and accountability as scientific workflows become increasingly automated.
\end{takeaway}

\section{Example: Legitimate AI Contribution vs. Misuse in Scientific Writing of Related Work}\label{sec:example}

To make the AI Research Object (AI-RO) framing concrete, we present a minimal paired example contrasting a legitimate use of generative AI in scientific writing with an opaque misuse scenario. Both cases involve preparing the related work section of a research paper. We focus on literature review writing because it is among the most common applications of AI in scientific authorship and is frequently associated with risks such as hallucinated citations and bias~\cite{ansari2026compounddeceptionelitepeer, vasu2025justice}. The distinction lies not in whether AI is used, but in how its role is bounded, documented, and integrated into the research workflow.

In the legitimate case, the author uses an LLM as an assistive synthesis tool while retaining full intellectual responsibility. The process begins with a curated set of verified references and human-authored summaries derived from direct reading. The model is prompted to organize these materials and suggest comparative phrasing, which is treated as provisional text rather than final output. The author verifies all claims and citations against primary sources and revises the draft to reflect the paper’s contribution.

Crucially, this interaction is preserved as an AI-RO artifact. Model identity, configuration context, and interaction traces are recorded, and human edits distinguish AI suggestions from final authorial decisions. The accompanying disclosure clarifies that the model supported structural synthesis and language refinement, while all scholarly judgments remained with the authors. Although stochastic generation limits exact regeneration, the artifact preserves a meaningful provenance trail that allows inspection of how AI influenced the workflow. Under this framing, the contribution is legitimate because the AI system operates as a transparent, bounded component of the research process.

The misuse case illustrates the opposite pattern: substitution of undocumented AI output for scholarly synthesis. Here, the author prompts the model to produce a complete related work section without supplying verified inputs or enforcing validation constraints. The resulting text may contain fabricated or misattributed citations, yet is incorporated with minimal review. Only a generic disclosure is provided, and no interaction records are retained. From an AI-RO perspective, the absence of provenance eliminates inspectability, making it impossible to reconstruct how claims were produced.

This paired contrast highlights a central claim of this paper: legitimacy in AI-assisted scientific writing is procedural rather than binary. Ethical use depends not on the presence of AI, but on whether its role is documented, bounded, and subject to the same expectations of transparency and accountability that govern other research artifacts.

\begin{takeaway}
    \textbf{Takeaway 4.} The paired example shows that legitimacy in AI-assisted scientific writing depends on procedural transparency and documented human oversight, not on the mere presence or absence of AI tools.
\end{takeaway}

\section{Demonstration: Inspectable AI Workflows}

\begin{figure}
    \centering
    \includegraphics[width=\linewidth]{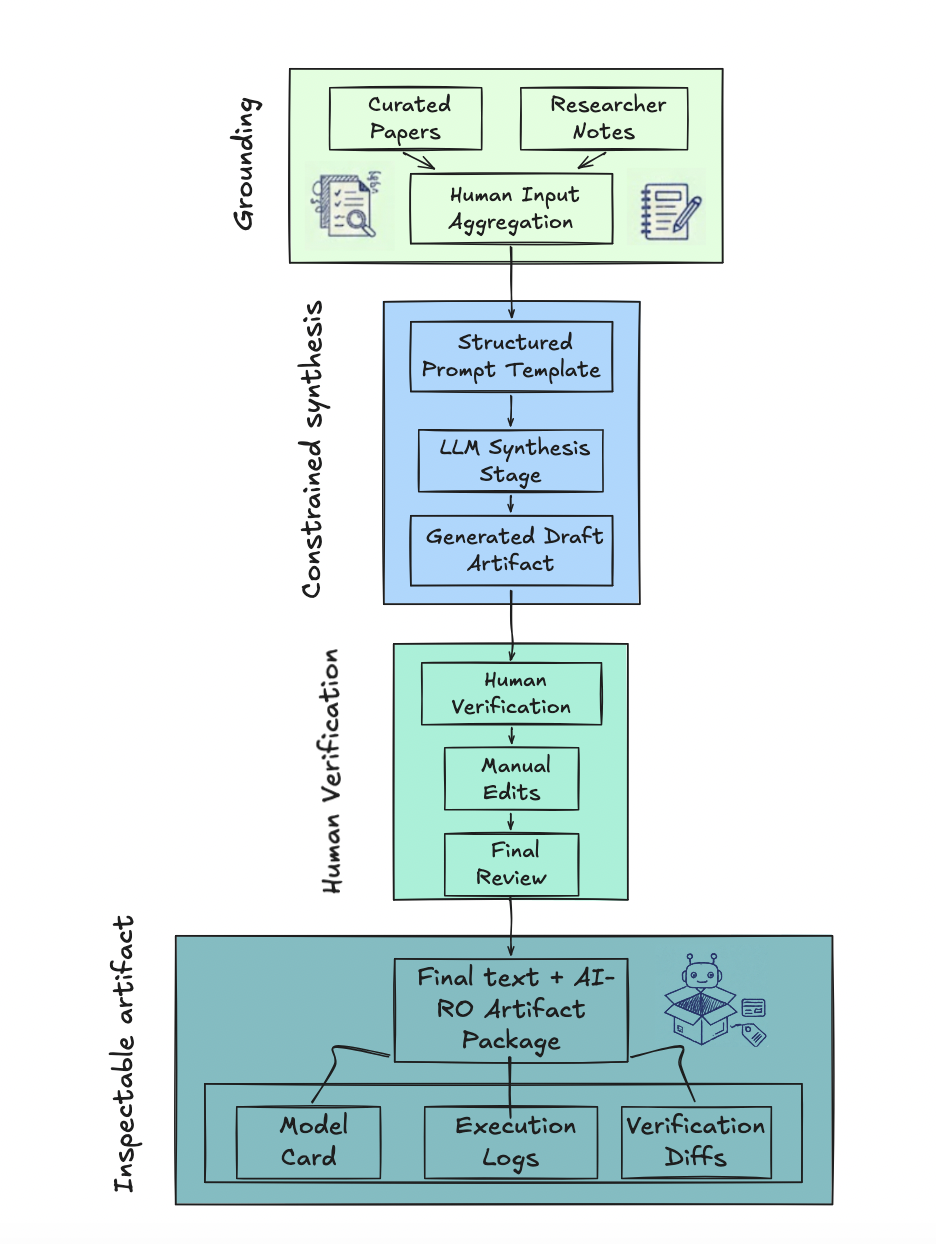}
    \caption{AI-RO Workflow for Literature Review Writing: From structured human input to verifiable outputs.}
    \label{fig:pipeline}
\end{figure}

To illustrate the practicality of the AI Research Object (AI-RO) framework, we present a workflow demonstration of an inspectable AI-assisted literature review pipeline (~\autoref{fig:pipeline}). The purpose of this demonstration is to illustrate how documentation practices shape the inspectability and governance of AI-assisted scholarship, in contrast to unstructured, opaque use.

\paragraph{\textbf{Workflow Design}}

 The example workflow implements an AI-RO  in which generative AI functions as a bounded synthesis tool embedded within a documented research pipeline. Human-authored inputs, including verified references, structured reading notes, and a contribution statement, are supplied to the model under explicit constraints. Outputs are treated as provisional drafts subject to human verification, and all model interactions are recorded as structured provenance artifacts.
A detailed procedural specification of the workflow is provided in the Appendix~\autoref{app:method}.

\paragraph{\textbf{Artifact Demonstration}}

Example workflow produces a concrete AI-RO artifact package that exposes the internal structure of AI-assisted synthesis. The central artifact is an AI Research Object inspection card (AI-ROIC) (Figure~\ref{fig:model-card}), which summarizes model identity, configuration parameters, released artifacts, intended use, human oversight, and reproducibility considerations. While traditional model cards describe a model’s capabilities, limitations, and recommended uses~\cite{mitchell2019model}, AI-ROIC functions as a compact inspection interface, allowing readers to understand how the generative system was configured, what materials are available for audit, and how responsibility is maintained. 

Supporting artifacts further expose workflow mechanics. Figure~\ref{fig:litreview-prompt} illustrates the prompt template used to bind model behavior to verified human-authored inputs. The artifacts and their role in auditing frameworks are summarized in~\autoref{tab:airo-artifacts}. The artifact contains an executable workflow, structured inputs, generated outputs, and provenance documentation packaged as an RO-Crate. To preserve double-blind review and avoid disclosure of manuscript text, interaction transcripts and timestamps are redacted and replaced with integrity hashes, while retaining configuration parameters and workflow structure. A detailed description of the artifact packaging, RO-Crate structure, and anonymization procedure (including log redaction and provenance preservation) is provided in Appendix~\ref{app:method}.

\paragraph{\textbf{Discussion}} 

While we present illustrative examples of AI-related research artifacts, we do not prescribe these templates as fixed standards. The structure and content of such artifacts should be adapted to the requirements of specific venues and workflows. For instance, the strictness of prompt templates may vary with the intended level of model autonomy: some templates may allow external references while requiring the model to produce verification checklists for human review. Regardless of configuration, it is advisable to place hard constraints at the beginning or end of the template, as LLMs tend to exhibit reduced attention to instructions embedded in the middle of longer prompts~\cite{liu2024lost}. Other directions may also include integrating retrieval-augmented generation (RAG) systems tailored for literature analysis and synthesis, enabling tighter coupling between curated sources and model outputs~\cite{asai2026synthesizing}.

The artifacts discussed here focus on a single component of the research lifecycle, namely, literature review, as a concrete demonstration. Comparable AI-RO pipelines could be developed for other stages, including ideation, experimental design, data analysis, and automated experimentation. Together, these components could form a larger, integrated AI-RO pipeline that covers the entire manuscript development lifecycle. Consequently, in more advanced settings, releasing AI agents that assist across multiple stages of research as a part of a documentation package may provide a unified transparency layer spanning the entire workflow.

\begin{table*}[t]
\centering
\footnotesize
\begin{tabular}{p{0.15\textwidth} p{0.30\textwidth} p{0.30\textwidth} p{0.12\textwidth}}
\hline
\textbf{Artifact} &
\textbf{Captured Information} &
\textbf{Governance Function} &
\textbf{Primary User} \\
\hline

AI-ROIC &
Model identity, configuration parameters, intended use, released artifacts, limitations, and oversight description &
Provides a high-level inspection interface summarizing how AI was used and governed; supports transparency and interpretability &
Authors, Readers, reviewers \\

Prompt template &
Constraints, task framing, allowed inputs, and generation rules &
Documents how model behavior was bounded; prevents hidden or unconstrained synthesis &
Authors,  reviewers \\

Interaction/invocation logs &
Timestamps, parameters, cryptographic hashes, prompt–response linkage &
Preserves provenance and enables auditability without exposing sensitive raw content &
Authors, Auditors, editors \\

Human-authored input bundle &
Verified references, structured notes, contribution framing &
Establishes human grounding of claims and prevents fabricated or unsupported synthesis &
Authors, reviewers \\

Generated draft artifact &
Model-produced synthesis text and intermediate reasoning structure &
Exposes how AI influenced writing; enables comparison with final authorial decisions &
Auditors, researchers \\

Human edit trace &
Tracked revisions distinguishing AI suggestions from author decisions &
Demonstrates human verification and responsibility over final claims &
Reviewers, auditors \\

RO-Crate metadata manifest &
Machine-readable description of files, relationships between artifacts, authorship roles, and provenance references &
Enables interoperability and automated inspection &
Repositories,  automated analysis systems \\

\hline
\end{tabular}
\caption{AI Research Object artifact stack and their governance roles. Each component of AI-RO externalizes a different aspect of AI-assisted workflow, enabling inspection, accountability, and procedural traceability.}
\label{tab:airo-artifacts}
\end{table*}

\begin{takeaway}
\textbf{Takeaway 5.}
By structuring literature review synthesis as an AI Research Object, the workflow demonstrates how provenance artifacts enable transparency, human oversight, and governance of generative AI.
\end{takeaway}

\begin{figure}[t]
\centering
\footnotesize


\begin{tcolorbox}[
  colback=blue!3,
  colframe=blue!60!black,
  title=\textbf{PROMPT TEMPLATE},
  fonttitle=\bfseries,
  boxrule=0.6pt,
  arc=1mm,
  width=\linewidth
]

You are assisting with drafting a RELATED WORK section for a scientific paper on  \{\{TITLE\}\}.\\

You will be given:
\begin{itemize}
    \item a topic
    \item a contribution statement
    \item a taxonomy (clusters and assignments)
    \item structured human notes for each paper (summary, strengths, limitations, relation)
\end{itemize}

Task:
Draft a related work section (target length: ~\{\{TARGET\_WORDS\}\} words) that:
\begin{itemize}
    \item Organizes discussion around the provided clusters
    \item Uses neutral, scholarly tone
    \item Highlights similarities and differences with respect to the contribution statement
    \item Mentions strengths/limitations carefully
    \item Avoids overclaiming (if evidence is weak, use cautious language)
\end{itemize}

Hard constraints (must follow):
\begin{itemize}
    \item Use ONLY the provided papers and identifiers. Do NOT invent citations or claims.
    \item Every comparative claim must be supported by the human notes provided.
    \item Use citations in the form: (\{\{citation\}\}; \{\{pid\}\}).
    \item If a claim is not supported by the notes, mark it as [NEEDS HUMAN CHECK].
\end{itemize}

Output format:
1) "RELATED WORK (DRAFT)" header
2) Draft text in paragraphs
3) "CLAIM CHECKLIST" bullet list with supporting paper IDs

\end{tcolorbox}

\caption{Prompt template used to guide the language model in drafting a structured literature review section under human-authored constraints.}
\label{fig:litreview-prompt}

\end{figure}

\section{From AI Research Objects to Governance Interfaces}

The AI-RO framework establishes how generative AI interactions can be documented as structured, inspectable components of a research workflow. A complementary question is how such artifacts might be surfaced within real scholarly processes, particularly peer review, without introducing excessive friction or compromising confidentiality. 

To operationalize this idea, we propose a conceptual governance interface in which AI-RO documentation is packaged as a companion artifact submitted alongside the manuscript. Rather than exposing full conversational logs, this interface surfaces curated provenance indicators, such as model identity, configuration summaries, interaction categories, and evidence of human verification, that communicate how generative assistance was integrated into the workflow. Access to these artifacts can be tiered according to venue policy, preserving confidentiality while enabling meaningful inspection.

This interface reframes AI disclosure from a declarative statement into a structured accountability layer. Reviewers gain visibility into the procedural role of AI systems without being required to audit raw transcripts, and editors can evaluate compliance using standardized artifact summaries. Crucially, the model emphasizes incremental integration: AI-RO artifacts are designed to complement existing submission infrastructures rather than replace them, minimizing additional friction for authors and reviewers.

In this way, governance becomes an extension of documented workflow provenance rather than an external compliance burden. Structured AI artifacts enable fine-grained transparency, support reproducibility, and align AI-assisted scholarship with familiar mechanisms of scientific accountability, while remaining practical for adoption within contemporary publishing pipelines.

\begin{takeaway}
\textbf{Takeaway 6.}
Standardized interfaces for accessing and analyzing AI-RO artifacts provide reviewers with actionable transparency while preserving confidentiality and minimizing friction.
\end{takeaway}

\section{Challenges \& Future Directions}

While the AI-RO framework provides a structured path toward transparency and accountability in AI-assisted scholarship, its practical adoption raises intertwined social, technical, and institutional challenges. Addressing these barriers is essential if inspectable AI workflows are to move from conceptual proposal to routine scholarly practice.

\paragraph{\textbf{Incentives for Transparency}}
Currently, disclosing AI assistance may negatively affect judgments of competence or credibility, creating incentives for underreporting~\cite{schilke2025transparency}. In environments where transparency is penalized, responsible documentation becomes risky. Overcoming this stigma requires shifting evaluation norms away from whether AI was used toward how it was integrated, verified, and governed. In addition, it is necessary to incentivize transparency by rewarding the submission of AI-RO and by simplifying the AI-RO generation and disclosure pipelines.

 \paragraph{\textbf{Security, Privacy, and Controlled Disclosure}}
AI-RO provenance artifacts may contain sensitive information, including unpublished findings, proprietary datasets, or 
vulnerability details that cannot be publicly disclosed. Beyond confidentiality, the artifacts themselves are potential attack surfaces: interaction logs or metadata could be modified 
post-hoc to misrepresent how AI was used, undermining the auditability the framework is designed to provide, while releasing structured provenance artifacts could inadvertently 
expose methodological details enabling adversarial replication.  These risks do not argue against provenance documentation, but call for treating AI-RO infrastructure with the same security discipline applied to research data and code. Concretely, they motivate mechanisms such as privacy-preserving logging, redaction with cryptographic hashing (as illustrated in our artifact design), and secure storage environments. In addition, S\&P community developed coordinated vulnerability disclosure to balance transparency with harm 
prevention. AI-RO governance can draw directly on this experience: rather than treating transparency as binary, the framework supports controlled disclosure, where provenance 
artifacts are released in tiers calibrated to research sensitivity and recipient trust level. A reviewer may receive configuration summaries and artifact hashes; a trusted auditor may receive redacted logs. This architecture mirrors responsible disclosure norms and may prove a more productive framing for the S\&P community than a requirement for full openness.

\paragraph{\textbf{Compliance Burden}}
Structured provenance introduces additional work for authors and new inspection responsibilities for editors and reviewers. Without automation and usable interfaces, AI-RO workflows risk becoming administratively heavy. Progress in automated metadata capture and lightweight audit tooling will be critical to ensuring that accountability mechanisms support, rather than obstruct, scholarly productivity.

\paragraph{\textbf{Ethical Considerations and Data Governance}}
Broader concerns surrounding copyrighted training data, dataset provenance, and systemic bias remain unresolved across the AI ecosystem. While the AI-RO framework does not directly resolve these issues, it provides a structured context for documenting and highlighting them. Future governance models may link transparency incentives to the use of ethically sourced or auditable AI models.

\paragraph{\textbf{Reproducibility}}
Reproducibility in generative systems is inherently limited. LLM outputs remain stochastic even under fixed seeds or low-temperature sampling~\cite{blackwell2024towards}, and proprietary models restrict access to internal representations. AI-RO practices, therefore, emphasize approximate reproducibility through preserved configuration, interaction context, and artifact hashes.  Namely, it allows reconstruction of the procedural context and regeneration of functionally similar outputs under equivalent configurations, rather than reproducing identical text. Further methodological standards will be required to determine what levels of variation remain epistemically acceptable in probabilistic systems.

\paragraph{\textbf{Scalability and Scope}}
 A practical challenge concerns the scale of AI-assisted interactions within real research workflows. Not all interactions are equally relevant to the final scientific contribution. AI-RO practices therefore require defining scope boundaries, capturing interactions that materially influence reported results while allowing ephemeral or exploratory uses to remain out of scope. Future systems should support multiple levels of granularity, ranging from coarse workflow summaries to fine-grained prompt-level logs, with increasing automation to reduce author burden. Such mechanisms are essential to ensure that AI-RO adoption remains practical for complex, iterative research processes.

\begin{takeaway}
\textbf{Takeaway 7.}
AI governance is not solely a technical problem, but a socio-technical transition that requires cultural norms, infrastructure design, and methodological innovation.
\end{takeaway}

\section{Conclusion, Limitations \& Call for Action}

Generative AI is rapidly becoming embedded in the everyday practice of scientific writing. Attempts to govern this shift through authorship debates, detection tools, or blanket prohibitions struggle to capture the infrastructural role that AI now plays in research workflows. This paper argues for a reframing: treating AI systems as \emph{Research Objects} whose configuration, interaction history, and outputs can be documented, inspected, and evaluated alongside data and software artifacts.

By operationalizing AI usage as structured provenance, the AI-RO framework shifts governance from attribution disputes toward methodological transparency. Beyond a governance proposal, AI-RO serves as a metascientific intervention in how epistemic provenance is recorded and evaluated. 
The paired workflow example and documentation demonstration illustrate that legitimacy in AI-assisted scholarship depends less on whether AI is used than on whether its influence is inspectable. The companion model interface example further shows how such documentation can be integrated into peer-review infrastructure without compromising confidentiality. 
 From a security and privacy perspective, the need for trustworthy and inspectable AI-assisted workflows is particularly urgent. AI-ROs provide a foundation for auditable, privacy-aware, and integrity-preserving research processes, enabling the scientific community to adopt generative AI without compromising core principles of accountability and reproducibility. We argue that future research infrastructures, particularly in S\&P domains, should incorporate mechanisms for secure AI-assisted workflows' provenance capture and controlled disclosure as first-class requirements.
\paragraph{\textbf{Limitations}} While we demonstrate the advantages and feasibility of the AI-RO approach through an example workflow, we acknowledge remaining limitations. The framework and demonstration intentionally focus on a single stage of the research lifecycle (i.e., drafting a RELATED WORK section) and do not constitute a comprehensive governance solution. While the proposed principles are designed to generalize beyond the literature review example, extending them to other stages of scientific workflows will require domain-specific adaptation, tooling, and validation.  In addition, the proposed workflow assumes standalone structured interaction with AI, whereas real-world AI use is often exploratory, iterative, and difficult to scope. Addressing this problem requires novel provenance tools that span multiple platforms and link fragmented interactions. Finally, a rigorous validation of the framework requires human study to evaluate the usefulness of the artifacts for scientific review and to assess the trustworthiness of the scientific work. Accordingly, the framework should be understood as an initial architectural proposal that delineates a design direction rather than a fully operational standard.
Finally, we point toward a future research agenda.  Below, we list the most urgent action points to support the use of transparent and ethical AI in scientific research.

\begin{action}
\small
\medskip

\cmark \hspace{5pt}\textit{Standardization of AI usage documentation} \newline

\cmark \hspace{5pt} \textit{Developing privacy-aware provenance infrastructure and anonymization protocols for AI-RO} \newline

\cmark \hspace{5pt} \textit{Supporting cultural norms that reward transparency usable audit interfaces} \newline

\cmark \hspace{5pt} \textit{Encouraging the development and adoption of open and responsibly trained models}\newline

 \cmark \hspace{5pt} \textit{Developing protocols and tools guiding AI use in research}\newline

 \cmark \hspace{5pt} \textit{Building tools and interfaces to support provenance in unstructured, iterative, and fragmented interactions with AI} \newline

 \cmark \hspace{5pt}\textit{Developing AI-RO governance norms tailored to security and privacy research, where confidentiality, integrity, and auditability of provenance artifacts carry domain-specific stakes}
\end{action}



  \section*{Acknowledgments}

The project is partially funded by the Deutsche Forschungsgemeinschaft (DFG, German Research Foundation), under grant agreement No. 568154038. This work is also funded by ELSA – European Lighthouse on Secure and Safe AI funded by the European Union under grant agreement No.101070617. Views and opinions expressed are however those of the author(s) only and do not necessarily reflect those of the European Union or the European Commission. Neither the European Union nor the European Commission can be held responsible for them. This work was also partially funded by the German Federal Ministry of Education and Research (BMBF) under the grant
AIgenCY (16KIS2012).



%



\bibliography{ref}
\bibliographystyle{abbrv}


\appendix~\label{app:method}

\section{Artifact Packaging and Demonstration Workflow}
\label{app}

This appendix describes how the AI-RO  corresponds to the literature review workflow example introduced in the paper and how it was prepared for anonymous inspection. 
The goal of the released artifact is to provide an inspectable execution trace of the proposed literature review workflow rather than to evaluate writing quality. The artifact, therefore, contains a representative run illustrating inputs, model invocation, verification steps, and provenance capture, and can be examined independently of the manuscript topic.

\subsection{Workflow Overview}

 AI-assisted writing is operationalized as a bounded synthesis process embedded in a human research pipeline. The workflow proceeds in five stages:

\begin{enumerate}
\item preparation of structured input notes,
\item constrained model invocation,
\item intermediate artifact generation,
\item human verification,
\item provenance recording.
\end{enumerate}

The language model functions as a synthesis and organization tool. 

\subsection{Structured Inputs}

The model is never prompted with open-ended literature queries. Instead it receives a curated input bundle consisting of human-authored reading notes. Each record follows a fixed schema:

\begin{quote}
identifier, citation, persistent identifier, summary, strengths, limitations, and relation to the contribution
\end{quote}

These notes are derived from direct reading. The prompt constrains the model to reorganize and summarize only this material. The model is explicitly instructed not to invent sources or claims.

\subsection{Model Invocation}

The model is invoked in two stages. First, the model groups papers into thematic clusters and provides a brief rationale for each cluster.

Second, the model produces a structured narrative outline derived only from the provided notes.

Hard constraints embedded in the prompt require that:
citations originate from the input bundle,
comparative statements correspond to human notes,
and uncertainty is explicitly flagged.

\subsection{Generated Artifacts}

Each run produces inspection artifacts rather than only a final text. The artifact includes:

\begin{itemize}
\item a taxonomy file describing conceptual groupings,
\item a demonstration draft showing synthesis structure,
\item an audit checklist linking claims to sources,
\item a verification table for human review,
\item a provenance record of model interaction.
\end{itemize}

These outputs externalize intermediate reasoning and make the writing process reviewable. The implementation is available at~\href{https://github.com/RutaBinkyte/AI-RO}{\texttt{https://github.com/RutaBinkyte/AI-RO}}. The research artifact is available at OSF.io:~\href{https://osf.io/zbrvu}{\texttt{https://osf.io/zbrvu}}.

\subsection{Provenance Logging}

Every model invocation generates a structured log containing:

\begin{itemize}
\item model identity and configuration,
\item generation parameters,
\item cryptographic hashes of prompt, response, and input bundle.
\end{itemize}

Hashes allow verification that an artifact corresponds to a specific input without exposing the underlying text. The released artifact therefore enables auditability while preventing disclosure of manuscript material.

\subsection{Log Redaction and Anonymization}

For double-blind review, interaction transcripts are not publicly distributed. Instead:

\begin{itemize}
\item full prompts and responses are replaced by integrity hashes,
\item timestamps are removed,
\item local file paths and machine identifiers are removed,
\item backend service endpoints are generalized.
\end{itemize}

This preserves verifiability while preventing attribution or reconstruction of the manuscript.

\subsection{RO-Crate Structure}

The artifact is packaged using the RO-Crate specification, which describes research objects using machine-readable metadata. The released archive contains:

\begin{itemize}
\item source code implementing the workflow,
\item prompt templates,
\item structured input notes,
\item redacted provenance logs,
\item generated demonstration outputs,
\item RO-Crate metadata describing file roles and relationships.
\end{itemize}

The file \texttt{ro-crate-metadata.json} declares each component as code, data, or provenance material and links outputs to the process that produced them. The package can therefore be inspected both manually and programmatically.

\subsection{Verification Procedure}

The audit workflow is designed so a reviewer can:

\begin{enumerate}
\item inspect structured notes,
\item examine generated clusters and draft structure,
\item check claim-to-source mappings,
\item verify artifact hashes,
\item confirm that outputs derive from the recorded inputs.
\end{enumerate}

Human verification is required before any generated text can be treated as scholarly writing.

\end{document}